\pdfoutput=1

\documentclass[11pt]{article}

\usepackage[final]{acl}
\usepackage{times}
\usepackage{latexsym}

\usepackage[T1]{fontenc}
\usepackage{booktabs}
\usepackage{multirow}
\usepackage{diagbox}
\usepackage{paralist}

\usepackage[utf8]{inputenc}

\usepackage{microtype}

\usepackage{inconsolata}

\usepackage{graphicx}
\usepackage{subcaption}
\usepackage{tabularx}
\usepackage{makecell}

\newcommand{\nnote}[1]{\textcolor{red}{$\ll$\textsf{#1 --Nir}$\gg$}}
\title{Finding your MUSE: Mining Unexpected Solutions Engine}

\usepackage{lipsum}

\newcommand{\remove}[1]{}

\author{
    Nir sweed$^{\spadesuit}$, Hanit Hakim$^{\spadesuit}$, Ben Wolfson$^{\diamondsuit}$, Hila Lifshitz$^{\triangle}$, Dafna Shahaf$^{\spadesuit}$ \\
    $\spadesuit$  The Hebrew University of Jerusalem \quad $\diamondsuit$ New York University \\
    $\triangle$ Warwick Business School \\
    \texttt{\{nir.sweed, hanit.hakim\}@mail.huji.ac.il} \quad \texttt{bw916@stern.nyu.edu} \quad \\ \texttt{hdiginnovation@gmail.com}  \quad \texttt{dshahaf@cs.huji.ac.il}
}
\date{}

\newcommand{\xhdr}[1]{\vspace{1mm}\noindent{{\bf #1.}}} 
\begin{document}
\maketitle

\begin{abstract}
Innovators often exhibit \emph{cognitive fixation} on existing solutions or nascent ideas, hindering the exploration of novel alternatives.
This paper introduces a methodology for constructing Functional Concept Graphs (FCGs), interconnected representations of functional elements that support abstraction, problem reframing, and analogical inspiration. Our approach yields large-scale, high-quality FCGs with explicit abstraction relations, overcoming limitations of prior work. We further present MUSE, an algorithm leveraging FCGs to generate creative inspirations for a given problem.
We demonstrate our method by computing an FCG on 500K patents, which we release for further research.
%
%
A user study indicates that participants exposed to MUSE's inspirations generated more creative ideas, both in terms of absolute number (up to 19\% increase over participants not given inspirations) and ratio (75\%, compared to 49\% for no inspirations).
\end{abstract}

\section{Introduction}
\label{sec: introduction}
A well-documented challenge in design and problem-solving is \emph{fixation}, wherein individuals become prematurely attached to a narrow set of familiar solutions or features, thereby impeding the generation of truly novel concepts \cite{jansson1991design, purcell1996design}. This cognitive inertia can significantly limit the effective exploration of the design space -- the abstract space of all possible solutions to a given problem. Navigating this complex space to identify diverse and high-quality solutions requires systematic approaches that encourage cognitive flexibility and the consideration of a broad range of alternatives.

Existing methodologies for ideation often fall short in robustly guiding designers out of fixation traps or in structuring the vastness of the design space in a functionally meaningful way. While keyword-based search can retrieve superficially related concepts, it often fails to uncover {\it deeper functional analogies} that are crucial for creative leaps \cite{holyoak1996mental}. There is a growing need for representations that capture the core functional essence of ideas, enabling a more principled exploration of potential solutions.

\begin{figure}[b!]
\centering
    \includegraphics[width=0.85\linewidth]{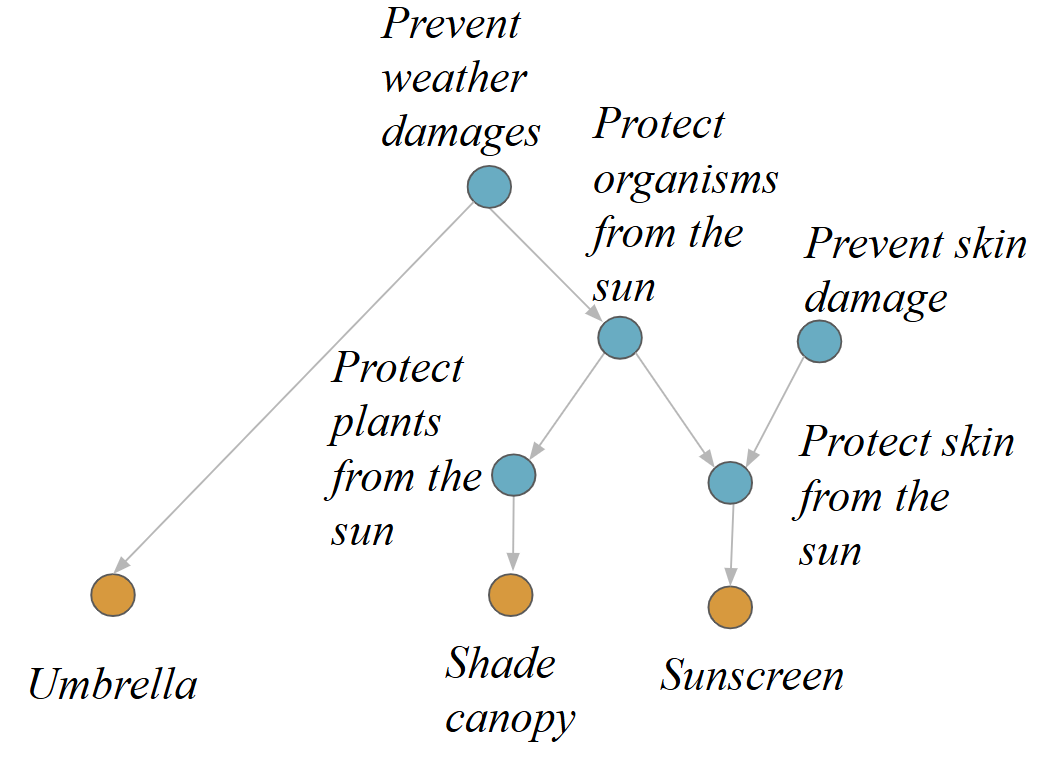}
    \caption{An example of a Functional Concept Graph. Node represent  problem  (spurposes, in blue) and solutions (mechanisms, in orange). Connections between problem nodes indicate abstraction. Connection between solution and problem nodes indicates the mechanism can solve the problem.
    }
    \label{fig:graph_example}
\end{figure}

We build upon an idea explored in recent work, that suggested decomposing ideas into their fundamental \emph{purposes} (problems) and \emph{mechanisms} (solutions) and organizing these into a structured representation called  a \emph{Functional Concept Graph} (FCG) \cite{accelerating_analogies_5}. In FCGs nodes correspond to purposes and mechanisms of products; edges either link between purposes and mechanisms that achieve them, or between purposes and their \emph{abstraction} (i.e., a more general problems).

We posit that such a representation can serve as a powerful cognitive scaffold for designers and problem-solvers. Specifically, by enabling navigation across interconnected functional elements, it facilitates the abstraction of problem statements and the discovery of \emph{analogical inspirations} from domains that might seem unrelated at the surface level \cite{gentner1983structure, gick1980analogical}.

For example, see Figure \ref{fig:graph_example}. Suppose an inventor wishes to protect their plants from the sun. They are familiar with standard mechanisms, such as  shade canopies. Through the graph, they might discover analogous mechanisms linked to the same abstract purpose. For example, they might reach the purpose node ``protect skin from sun'' through shared parent ``protect organisms from the sun'', which might inspire them to think of sunscreen for plants (which, surprisingly, has already been invented). Alternatively, they might explore even higher-level abstractions. This structured exploration can systematically help designers break free from initial conceptual ruts, reframe their understanding of the problem, and ultimately chart more innovative trajectories within the design space.

However, the implementation of \citet{accelerating_analogies_5} was very limited. Most notably,  their
edges only capture simple co-occurrence patterns in the corpus; there is no guarantee of explicitly encoding abstraction. In addition, their annotation process relied heavily on crowdworkers, which resulted in a noisy graph, and graph-building did not scale. 

In this work, we take advantage of the tremendous recent progress of LLMs and reimagine Functional Concept Graphs. Our contributions are:



\begin{compactitem}
\item We propose a novel, scalable approach to constructing Functional Concept Graphs that results in  richer, better-connected and less noisy graphs, whose edges explicitly encode abstraction relations. 
\item 
We compute an FCG on 500K patents, and release it for further research. 
\item We introduce MUSE, an algorithm that, given an FCG and a target problem, can produce inspirations for creatively solving the problem.
\item We conduct a user study and show MUSE inspirations can enhance human creativity. Our analysis shows that using our inspirations, participants were able to come up with up to 19\% more creative solutions, in comparison to participants that did not receive inspirations. More importantly,  75\% of the solutions produced by participants exposed to MUSE's inspirations were deemed creative (compared to 49\% for participants who did not see the inspirations).  
\item We release all the data and code used for creating the graph and analysis. \footnote{Code and data: \url{https://github.com/NSweed/MUSE}}
\end{compactitem}

\section{Functional Concept Graphs}
\label{sec:task}

Our goal is to automatically build a Functional Concept Graph (FCG) \cite{accelerating_analogies_5} and develop an algorithm to sample inspirations from it, given a target problem. 

Building upon the foundations of functional modeling, an FCG provides a structured graphical representation. Nodes embody functional concepts, encompassing both the intended purposes (problems tackled) and the underlying mechanisms (solutions) that enable these purposes. A directed edge between a purpose node and a mechanism node indicates that the mechanism is useful for achieving the purpose; a directed edge between two purposes indicates that the first is an abstraction of the second (see Figure \ref{fig:graph_example}).

\section{Data}
\label{sec:data}

We chose to test our idea on a patent corpus because it is vast, publicly available and contains a variety of real-life problems and solutions. In contrast, \citet{accelerating_analogies_5} has used a much smaller dataset of crowdsourced innovations.

We use a dataset taken from \href{https://patentsview.org/}{Patentsview.org} website \cite{toole2021patentsview},  
which contains patents registered in the U.S.~since the 1940s.
From each patent, we take its title and abstract text, which contains an informative description of the product. In addition, as part of our annotations, we used each patent \textbf{CPC} tag. 


\textbf{CPC} (Cooperative Patent Classification) is a hierarchical patent classification system, developed by the European and US Patent Offices. The system assigns each patent with (potentially multiple) CPC tags. 
Each tag has an id and a short name. 


Out of the full patent corpus extracted from the website, we included only patents that belong to 3 CPC top-level tags (out of 9 overall): Human Necessities (A), Operations and Transport (B), and Mechanical Engineering (F). We included those tags as they are likely to describe relatively simple everyday products, as opposed to sections that describe more specific professional domains such as Chemistry and Metallurgy (C). After the filtering, we were left with about 3M patents. Due to a budget limitations, we sampled 500K patents and used them as our corpus. 
\begin{figure*}[t]
\includegraphics[width=0.9\textwidth]{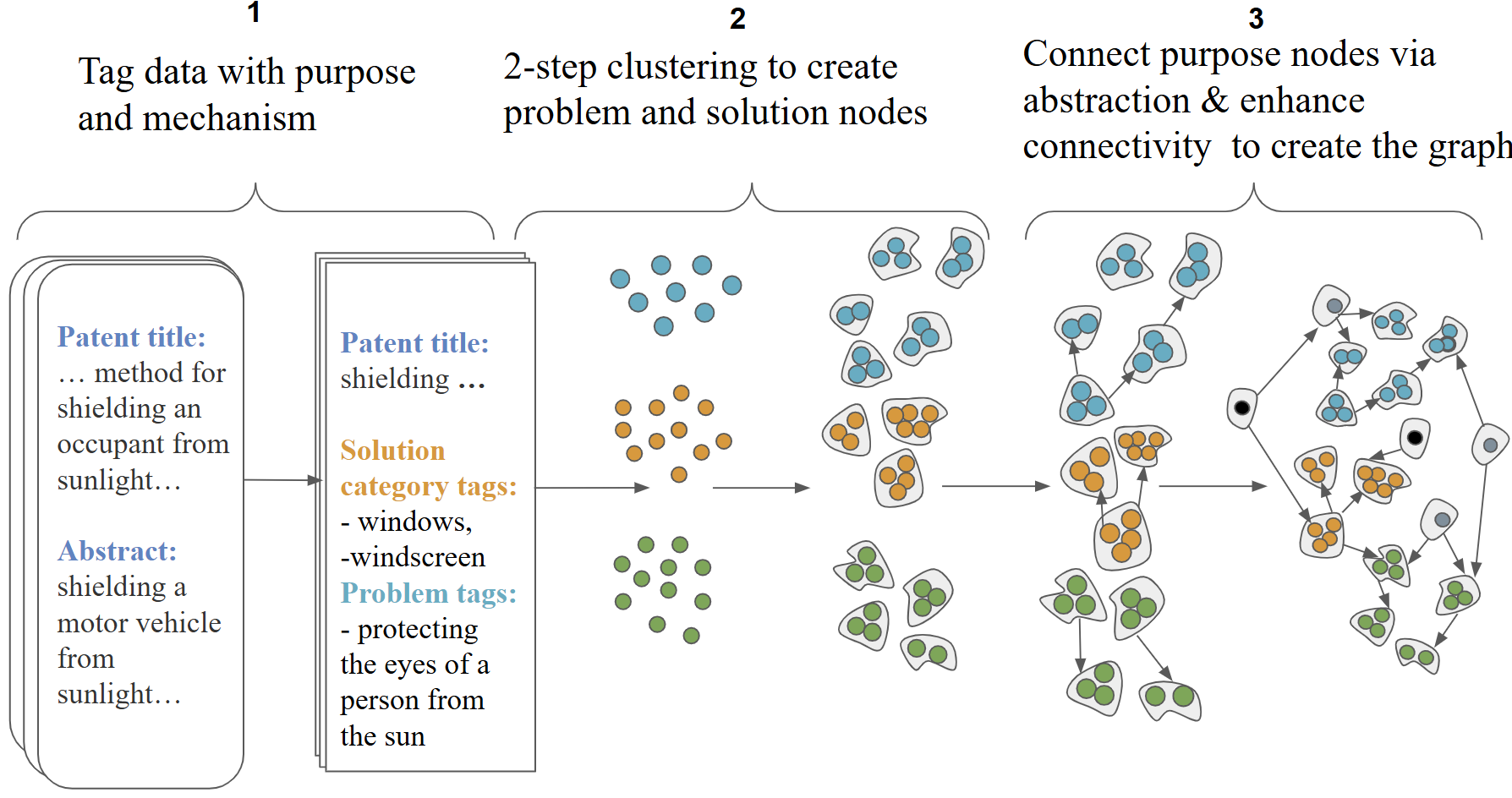}
\centering
\caption{A visualization of our full pipeline.  (1) We start by extracting purpose and mechanism tags from patent descriptions. (2) Then, we create the problem and solution nodes. To reduce the amount of computation, we first cluster the purpose tags to create loose clusters, then aggressively cluster each loose cluster to obtain the problem nodes.  This clustering induces the solution nodes as well. (3) The final step is to create edges connecting the problem nodes by finding abstraction relations, and then enhacing connectivity through virtual nodes.}

\label{fig:full pipeline}
\end{figure*}

\section{Constructing an FCG}
\label{sec:methods}



Our pipeline consists of 3 main steps (see Figure~\ref{fig:full pipeline}): (1) extracting  problems and solutions from a large dataset of patents, (2) creating the nodes of the FCG and (3) adding in the edges. 

\label{sec:Inspiration Graph: implementation}

\subsection{Extracting problems and solutions}\label{sec:Dataset:Annotations}
\label{sec:methods:getting annotations}

We start by annotating each patent with its \emph{purpose} (problem, what is it used for) and \emph{mechanism} (solution, how it works). 
We acquire multiple mechanism and purpose tags for each patent, as opposed to a single aggregated tag.

Importantly, patents are written in technical, legal language (``Legalese''). Surprisingly often, they are missing important commonsense information (for instance, a patent about airbags that never mentions cars or accidents, but rather focuses on the technical aspects of the invention). 
Thus, we take advantage of large-scale language models as well as patent metadata to annotate the dataset.




\xhdr{Getting mechanisms}
To extract mechanism tags for each patent, we make use of the CPC tags, that offer a granular breakdown of the technical features of the invention. 

We processed the full set of CPC categories (over 250,000 categories). We discard all CPC tags in the lowest level of hierarchy level, since they are too specific for our needs. We clean and preprocess the titles (see Appendix~\ref{sec:appendix:implementation details: mechanism}), and end up with a set of 8500 CPC tags.


Many CPC tags describe mechanisms, but not all. Thus, 
we manually tagged 1500 CPC tags as either related to mechanism or not. We used this annotated dataset to fine-tune a simple RoBERTa-based binary classification model \cite{liu2019roberta}. 
We train the model for 500 epochs, using 1e-7 learning rate. The final model we used achieved an F1 score of 0.88. 

\xhdr{Getting purposes}
Driven by its few-shot capabilities, we use GPT3 \cite{brown2020language} to generate \emph{purpose} tags. 
We adopt an in-context learning approach, and use the method proposed in \citet{prompt_engineering} to construct a prompt with 3 patent descriptions and their purpose tag annotations, followed by the patent description to be tagged (see Figure~\ref{fig:full pipeline} for example tags and Appendix~\ref{sec:appendix: annotation prompt} for prompt example). 
Although newer and more advanced models are available, we found that GPT3 (Babbage) provided the best performance-cost tradeoff for tagging 500K patents. 

\subsection{Creating problem and solution nodes}
\label{sec:methods: creating nodes}
After obtaining the purpose and mechanism tags, we move on to creating the problem and solution nodes that serve as our basic building blocks for the graph. 

\xhdr{Creating purpose nodes}
\label{sec:method:purpose nodes}
Our goal is to create problem nodes that cluster together conceptually similar purpose tags. 
To achieve this goal, we choose to use {agglomerative clustering} \cite{ward1963hierarchical} over {Sentence-BERT}  \cite{reimers2019sentence}  embeddings of the purpose tags. Agglomerative clustering allows us to control the similarity threshold, and also does not require specifying the number of clusters in advance. 

However, running agglomerative clustering over all purpose tags is costly.
We follow a common practice, where coarse‑partitioning strategy is used initially to break the data into manageable chunks, and a refinement clustering is performed within each chunk, significantly reducing the computation \cite{ma2018two}.
We use K-means \cite{macqueen1967some} to split the tags into loose clusters, and then run agglomerative clustering with a strict similarity threshold on each of them, 
making sure to keep semantically similar tags in the same cluster. 

We use the resulting clusters as the problem nodes in our graph. To select the parameters for the agglomerative clustering step, we generate a small dataset of 30 sentences (that did not appear in our data) and manually split them into clusters. We run the agglomerative clustering algorithm with similarity threshold ranging from 0.05 to 0.4, with increment of 0.05. We finally choose similarity threshold = 0.2, since it showed the best tradeoff between the purity (1.0) and NMI (0.97) metrics. See Appendix  \ref{sec:appendix:implementation details: clustering parameters} for full implementation details and hyperparameter selection.


\xhdr{Creating solution nodes}
We wish to create clusters of solutions used to solve similar problems. These solutions are not necessarily semantically similar themselves. Therefore, we do not cluster the mechanism tags directly, but rather induce the solution clusters from the problem clusters we created. 
We cluster two mechanism tags together if there exist purpose tags extracted from their corresponding patents that were clustered together in the previous step. 

\subsection{Adding edges}
\label{sec: methods: building graph}
The final step of creating the FCG is to connect the problem, solution nodes we obtained in section~\ref{sec:methods: creating nodes}.


As described in section~\ref{sec:task}, edges in FCGs reflect either an abstraction relation between problem nodes or a problem-solution relation.

\xhdr{Problem to solution} We connect a problem node  to a solution node 
if there is at least one patent that was assigned a problem tag in the problem node (i.e., in the corresponding cluster) and a solution tag in the solution node.



\xhdr{Problem to abstract problem}
Intuitively, if a problem (``protecting an organism'') is more abstract than another (``protecting plants''), the specific problem \emph{entails} the more abstract one. Hence, we use a pretrained NLI model \cite{laurer2024less} to identify entailment. As checking for entailment over all problem nodes is computationally expensive, we run the model over all pairs of problem nodes from the same loose cluster (Section \ref{sec:method:purpose nodes}). For each problem node, we select a representative purpose tag and add a prefix to turn it into a sentence (so it resembles the data the NLI model was trained on). We add an edge if the entailment score is above threshold $t$. See Appendix \ref{sec:appendix:implementation details: entailment} for details about entailment thresholds and prefixes. 

We note that the graphs formed by this process might include cycles, due to mistake or inconsistencies of the NLI model. Moreover, these graphs might also include redundant edges (if the graph contains edges $(x_1,x_2), (x_2,x_3)$, edge $(x_1,x_3)$ is redundant).  Therefore, we adopt the method suggested in \citet{sun2017breaking}  to remove cycles while maintaining the abstraction hierarchy. Then, we eliminate redundant edges in the graph by keeping only the longest paths between any connected nodes. The result of this process are $K$ interim graphs 
$G_1,...,G_K$ (one for each $K$-mean cluster).

\xhdr{Enhancing connectivity}
 Some abstract relations might not be captured by the NLI model. Thus, our algorithm creates two types of \emph{virtual} nodes to enhance the connectivity of problem nodes:

\begin{compactitem}
    \item \textbf{LLM-based connections.} We seek abstraction relations that were not captured by the NLI model. For each loose-cluster graph $G_i$, we automatically extract a set of candidate nodes $N_i$ that correspond to relatively general problems; the algorithm selects candidate nodes based on their height in the cluster and in their path (see Appendix \ref{sec:appendix: implementation details: LLMs} for details). We use llama3.1-8b-Instruct \cite{grattafiori2024llama} to suggest abstractions that capture several of these nodes. We create a virtual LLM-node for each  abstraction, and connect it with the nodes responsible for its creation.
    \item \textbf{Verb-based connections.} To capture far analogies in the graph, we choose to abstract problem nodes by the \emph{verb} appearing in them (e.g., protect).  We collect lists of synonymous verbs from WordNet \cite{fellbaum2010wordnet}, and create a virtual verb-node for each list of synonyms. Then, we connect  each node in the graph to all verb-nodes containing verbs appearing in it.
\end{compactitem}

The final step is to try to explicitly improve the connections between the interim graphs $G_1,...,G_K$, to allow the user to find \emph{far analogies}. We repeat the process of enhancing connectivity, but this time we focus on pairs of nodes coming from \emph{different} sets of candidate nodes,  $u\in N_i, v\in N_j, i\neq j$. See Appendix~\ref{sec:appendix: implementation details: interim graph}.

\section{MUSE: Sampling inspirations}
\label{sec: experimental setup: sampling}
Given a problem $p$, we wish to sample inspirations from the graph. To do so, we first encode the text describing the problem using Sentence-BERT, and then find the closest node to it in the graph $n_p$ using Faiss-index \cite{douze2024faiss}. After finding the anchor node, the next step is to sample inspiration nodes relevant to this node. 
%
The graph provides an intuitive interpretation of paths between nodes:  
For example, one could move up (to a more abstract problem) and then down (to a less abstract problem), reaching a sibling node (connected by a shared parent abstraction).  
Although many types of paths might produce useful inspirations,  we limit the paths from which we sample inspirations to focus on the classical analogy-making schema (``v-structure''), by allowing 1-2 abstraction steps, and a single concretion step (``up, down'' and ``up, up, down''). Exploring the effect of other types of paths is left for future work. 


\xhdr{Sampling nodes from different sources}   In  Section \ref{sec: methods: building graph} we described 3 types of connections between purpose nodes: NLI-based, verb-based and LLM-based. We define \emph{LLM-nodes} (\emph{verb-nodes}) as nodes for which at least one of the edges along the path connecting then with $n_p$ is LLM-based (verb-based). NLI-nodes are nodes for which all of the edges along the path are NLI-based. 

\xhdr{Ensuring diversity} Consider all nodes reachable from the start node $n_p$ by following paths (``up'', ``down'') or (``up'', ``up'', ``down''). To ensure the sampled nodes are both relevant and diverse, we use MMR \cite{carbonell1998use}  to select up to 5 nodes 
from each path and source (up to 30 nodes total). 
We refer to this sampling process as MUSE -- an algorithm aimed at helping users come up with novel ideas.

\section{Evaluation}
\label{sec: experimental setup}

Now that we have built our graph, 
our goal is to use it to help users find creative and novel solutions to their problems.  Specifically, we are interested in the following research questions:\\
\noindent \textbf{RQ1:} Can inspirations from the FCG enhance users' ability to come up with original solutions to problems? If so, what is the best way to communicate the inspirations to users?\\
\noindent \textbf{RQ2:} Which type of trajectories in the FCG produces more helpful inspirations?

\subsection{Experiment design}
\label{sec: exprimental setup: experiment design}
To answer these questions, we conducted a user study. 
Participants were randomly assigned a problem. 
After reading the instructions, they were given 15 minutes to come up with as many creative ideas as they could. 
Each participant was randomly assigned one out of 4 conditions, corresponding to different ways to display the inspirations (see below). In 3 of the conditions participants received inspirations drawn from the FCG; in the last condition no inspirations were given. 
The subjects that received inspirations were instructed to identify the source of inspiration for each ideas (which could be their own idea or one of the inspirations provided). The full instructions for the experiment are provided in Appendix~\ref{sec:appendix:expriment instructions}.

The experiment was carried out on the Prolific platform \cite{Prolific}. Participants were paid \textit{\pounds}3.25. 61 native English-speakers took part in the experiment, split almost evenly between conditions (16-15-15-15) and problems (31-30). 53.3\%  identified as females, and 46.7\% as males. 21.4\%  were in the 18-29 age group, while the percentages for the age groups 30-44, 45-59 and 60+ were 42.9, 28.6 and 7.1, respectively. 

\xhdr{Problems} We chose 2 everyday problems: ``Seal a leak'' and ``Cool a room''. The problems were selected from a list of everyday household problems from \href{https://www.ehow.com/}{ehow.com}. We chose these problems since they are very familiar to the common person, have well-known existing solutions while still enabling creative solutions. 


\xhdr{Displaying the inspirations}  Inspirations are problems sampled from the graph as described in section \ref{sec: experimental setup: sampling}. We define 4 conditions (ways to display them):
\begin{compactenum}
    \item \textbf{Purpose}: Showing just the purposes.
    \item \textbf{Purpose + mechanism}: Showing purposes  + up to 3 solutions sampled for each purpose.
    \item \textbf{Purpose + mechanism sentence}: Same as condition 2, but we use an LLM (Claude 3.7 Sonnet \cite{Claude37sonnet}) to turn inspirations into full sentences.
    \item \textbf{Empty} No inspirations.
\end{compactenum}
One example of a problem sampled as inspiration for the problem ``Cool a room'' is \textit{``Creating a chamber seal mechanism''}. A solution sampled for this problem is \textit{``Packing rings''}, and the generated sentence  is  {``Packing rings expand under fluid pressure to create effective chamber seals''}.
Additional examples are given in Appendix~\ref{sec: appendix: inspiration examples}. 

\begin{table}[t!]
  \centering
  \setlength{\tabcolsep}{3pt}
  \begin{tabular}{|l|r|r|r|r|}
    \hline
    & \textbf{Purp} & 
    \begin{tabular}[c]{@{}c@{}}\textbf{Purp+}\\\textbf{mech}\end{tabular} & 
    \begin{tabular}[c]{@{}c@{}}\textbf{Purp+}\\\textbf{mech}\\\textbf{sent}\end{tabular} & 
    \textbf{Empty} \\
    \hline
    Feasible (\#) & 4.37 & 3.67 & \textbf{4.8} & \textbf{4.8} \\
    Feasible (\%) & 0.72 & 0.78 & \textbf{0.83} & 0.63 \\
    \hline
    Creative (k=2) (\#) & 3.18 & 2.6 & \textbf{4.06} & 3.4 \\
    Creative (k=2) (\%) & 0.58 & 0.57 & \textbf{0.75} & 0.49 \\
    \hline
    Novel(k=3) (\#) & 1.81 & 1.13 & \textbf{1.86} & 1.73 \\
    Novel(k=3) (\%) & 0.31 & 0.26 & \textbf{0.32} & 0.25 \\
    \hline
  \end{tabular}
  \caption{For each condition, we report the total number and ratio (from all solutions) of feasible and creative solutions produced. We report both the liberal (novelty threshold $k=2$) and strict ($k=3$) settings. Although the empty condition produced the most feasible solutions, other conditions, especially the sentence condition, produced more novel solutions. All inspiration-based conditions produced a higher ratio  of feasible and creative solutions than the empty condition. }
  \label{tab:feasiblity novelty}
\end{table}

\section{Results}
\label{sec: results}
\label{sec: results: annotation}
We are interested in the degree to which our approach helps users to come up with creative solutions to well-known problems. For that, we measured the quality of the solutions produced by the participants in the experiment. Expanding upon \citet{reinig2007measurement, accelerating_analogies_3}, we define creativity as a combination of {\bf utility} and {\bf novelty}. Thus, we first score each solution with a binary feasibility score representing whether this solution is feasible and solves the given problem. Feasible solutions were then tagged with a novelty score (1: the solution is well-known, 2: uncommon solution, 3: a very novel solution).  In the following analysis, we treat a solution as either \emph{creative} or \emph{not creative}, by applying a novelty threshold $k$. We report results for both a liberal setting ($k = 2$) and a strict setting ($k = 3$). 

Overall, 3 judges tagged 374 solutions suggested by the participants. Since this is a non-trivial annotation task, the judges were first calibrated over 10 randomly selected solutions. Then, their agreement was computed over another set of 10 solutions. The remaining 354 solutions were randomly split into 3 and annotated separately by the judges. Agreement between the judges was substantial, with 90\% full agreement (all judges agreed) on the feasibility scores. Agreement for the novelty score was high as well, with 66\% agreement for the liberal case, and 88\% agreement for the strict case. For all scores and cases, agreement between at least 2 out of the 3 judges was 100\%.

\begin{table}[t!]
  \centering
  \setlength{\tabcolsep}{3pt}
  \begin{tabular}{|l|r|r|r|}
    \hline
    \backslashbox[3cm]{Metric}{Condition} & \textbf{Purpose} & 
    \begin{tabular}[c]{@{}c@{}}\textbf{Purp}\\\textbf{+mech}\end{tabular} & 
    \begin{tabular}[c]{@{}c@{}}\textbf{Purp}\\\textbf{+mech}\\\textbf{sent}\end{tabular} \\
    \hline
    \makecell{\% creative (k=2) \\\ from inspired} & 0.6 &  0.57 &  0.73 \\
    \hline
    \makecell{\% creative (k=2) \\ from non-inspired} &  0.44 &  0.47 &  0.47 \\
    \hline
    \hline
     \makecell{\% creative (k=3) \\ from inspired} &  0.42 &  0.25 &  0.37 \\
    \hline
     \makecell{\% creative (k=3) \\ from non-inspired} &  0.17 &  0.2 &  0.08 \\
    \hline
  \end{tabular}
  \caption{Percentage of creative solutions from inspired and non-inspired (=participant indicated this was their own idea) solutions. For all conditions and novelty thresholds, the percentage of creative solutions from inspired-solutions is noticeably higher.}
  \label{tab:inspired vs non}
\end{table}

\subsection{RQ1: Effect of inspirations}
\label{sec: results: RQ1}
We assess the degree to which each of the inspiration-based conditions (purpose, purpose + mechanism, purpose + mechanism sentence) increased the creativity of the participants. As can be seen in Table \ref{tab:feasiblity novelty}, participants in the empty condition (as well as the sentence condition) provided the highest absolute number of feasible solutions, perhaps because they did not spend time reading inspirations. However, the \emph{ratio} of feasible ideas (out of all ideas, averaged over all participants) was higher for the inspiration-based conditions, suggesting that {\bf inspirations helped participants produce higher-quality ideas}. 

For creativity, we see that for both novelty thresholds $k=2, k=3$, the purpose+mechanism sentence condition produced the highest number of creative ideas, but participants in all inspiration-based conditions produced a significantly higher \emph{ratio} of creative ideas compared to participants in the empty condition, again indicating the contribution of the inspirations.

We were surprised to see that the performance of purpose+mechanism was low compared to other inspiration-based conditions. This might indicate that the relation between mechanisms and purposes is not always clear,  confusing the participants and increasing the cognitive load, and putting it into a sentence helps participants. 

Since {\bf the sentence condition has yielded the best results}, we focus on it, and test its usefulness in producing creative ideas. We run a statistical test to compare the ratios of creative ideas per participant under the empty and sentence conditions. We verify normality and equal variances using the Shapiro-Wilk test and Levene's test, and run Student's t-test to compare the two conditions. The results for the liberal case are significant with $p=$ 0.004, but the results for the strict case were not ($p$ = 0.07), potentially because the number of highly novel ideas was smaller.



For the inspiration-based conditions, we examine the reported source of inspiration per idea. The ratio of creative solutions out of solutions inspired by the graph was higher than the ratio of creative solutions out of non-inspired ones (participants' own ideas) 
across all conditions (Table \ref{tab:inspired vs non}).

\subsection{RQ2: Trajectories}
\label{sec: results: RQ2} To answer our second research question, we compare the usefulness of the inspirations by their trajectory (NLI, LLM, Verb nodes). 

We look at the ratio of feasible and creative solutions out of all solutions inspired by a certain trajectory type. The results in Table \ref{tab:sources comparison 2} indicate that both LLM and NLI-based inspirations were able to produce solutions of higher quality than  verb-based inspired solutions. This might make sense, as two nodes that only share synonymous verbs might be very far off. To complement this, we observe the percent of inspirations from each source that were used in creative solutions. We find that 38\% of the verb-based inspirations were used in very creative solutions (novelty threshold = 3), close to that of the other two sources (44\%, 42\%). We conclude that {\bf although the LLM and NLI-based inspirations proved superior in our experiment, verb-based inspirations are still useful}. 

\begin{table}[t!]
  \centering
  \setlength{\tabcolsep}{3pt}

  \begin{tabular}{|l|r|r|r|}
    \hline
    \diagbox[height=2.5\line]{Metric}{Source} & \textbf{NLI} & 
    \begin{tabular}[c]{@{}c@{}}\textbf{LLM}\end{tabular} & 
    \begin{tabular}[c]{@{}c@{}}\textbf{Verb}\end{tabular} \\
    \hline
    \% Feasible solutions  & 0.84 & 0.78   & 0.73  \\
    \hline
    \% Creative solutions (k=2) &  0.71 &  0.67 &  0.54 \\
    \hline
    \% Creative solutions (k=3) &  0.39 & 0.4 &   0.29 \\
    \hline
  \end{tabular}%
  \caption{Percentages of feasible and creative solutions per trajectory type (out of all solutions using this trajectory). 
  The results are calculated over both problems from the experiment. For all metrics and cases, the results for NLI and LLM-based inspirations are higher than those of the verb-based inspirations, indicating a stronger signal for enhancing creative ideation.}
  \label{tab:sources comparison 2}
\end{table}

\subsection{Additional insights}
\label{sec:results:insights}

\begin{figure}[t]
    \centering
    \begin{subfigure}[b]{0.5\textwidth}
        \centering
        \includegraphics[width=\textwidth]{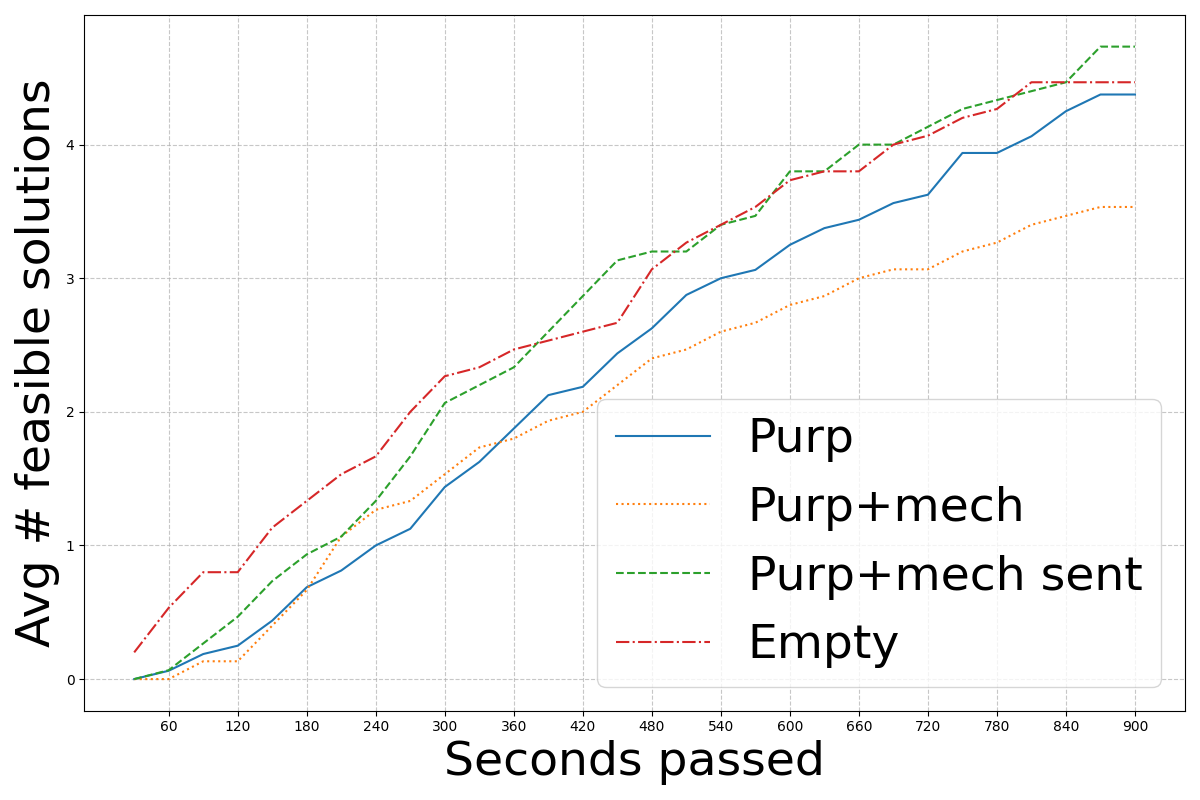}
        \label{fig:subfig1}
    \end{subfigure}
    \hfill
    \begin{subfigure}[b]{0.5\textwidth}
        \centering
        \includegraphics[width=\textwidth]{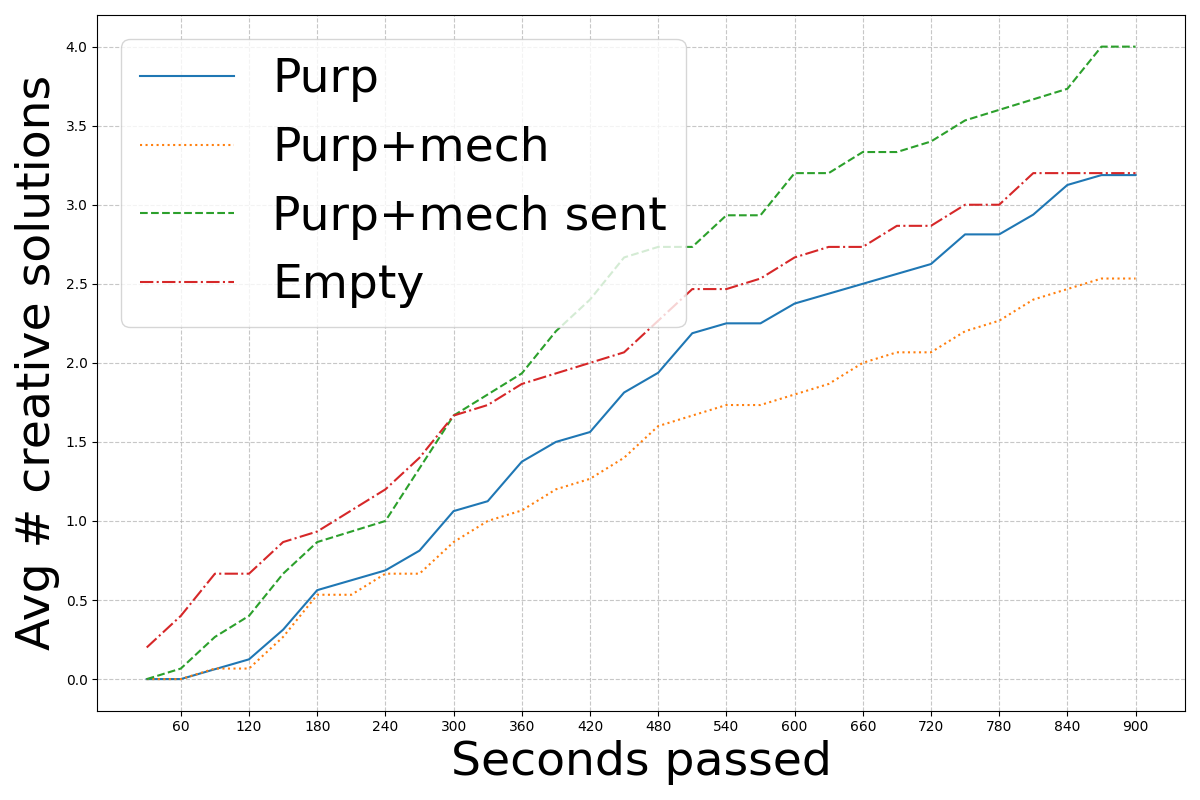}
        \label{fig:subfig2}
    \end{subfigure}
    \caption{Average number of feasible (top) and novel (bottom) solutions through time passed in the experiment for the different conditions. Participants under the empty condition (green dash-dotted line) produced more feasible and novel ideas at the start of the experiment. As time progressed, participants under the purpose (solid blue) and purpose+mechanism sentence (dashed green) conditions managed to come up with more feasible and novel ideas. This aligns with our hypothesis that generating solutions under the inspiration-based conditions requires additional time.}
    \label{fig:time analysis}
\end{figure}

\xhdr{Solution generation over time} We compare the number of feasible and creative solutions produced by participants under each condition, as a function of the time from the start of the experiment (Figure \ref{fig:time analysis}). Participants under the empty condition produced more feasible (top figure) and creative ideas (the figure depicts the liberal novelty setting, but this is true for the strict setting as well). We hypothesized that participants in the empty condition are not shown inspirations and can immediately start coming up with ideas, leading to better results at the beginning of the experiment. However, this advantage disappears over time, as the subjects in the other conditions exceed the performance of the empty-condition participants. We also note that 33.4\% of the participants under the empty condition stated they needed more time to complete the task, as opposed to 50\%, 46.7\% and 53.3\% of the participants under the inspiration-based conditions.

\xhdr{Preliminary exploration of SOTA LLMs}
Despite their tremendous popularity, state-of-the-art LLMs struggle with creative thinking and problem solving \cite{tian2023macgyver, franceschelli2024creativity}. We perform a preliminary study to assess the possibility of using LLMs to find creative solutions to everyday problems. We ask GPT-4o \cite{hurst2024gpt}, a popular SOTA LLM,  to generate original solutions to the same problems given to the participants in our experiment. We use similar instructions to those given to participants. Overall, GPT-4o produced 14 solutions (7 for each problem), 8 of which deemed feasible by our annotator. However, further analysis of the solutions showed that \emph{all} solutions already appear online, and some are common, 
hinting that SOTA LLMs might indeed be limited in producing truly novel solutions.

\xhdr{Satisfaction} After completing the task, participants were asked to completed a short survey. When asked whether similar inspirations would be helpful in the future when tackling a problem, 100\% of the participants under the sentence condition answered positively, compared to 75\% and 86.7\% of participants under the purpose and purpose+mechanism conditions. 



\section{Related Work}
\label{sec: related work: accelerating innnovation}

Computational methods aimed at augmenting human creativity and ideation have garnered significant attention. A prominent avenue  focuses on leveraging analogy as a powerful cognitive mechanism for generating novel solutions \cite{gentner1983structure, hofstadter2001analogy}. 
%
One significant line of work involves the creation and utilization of structured knowledge to identify potential analogies. For instance, the seminal Structure-Mapping Engine (SME) \cite{falkenhainer1989structure}
operates on propositional representations.


Harnessing analogies to navigate between ideas has been explored in  design-by-analogy works. 
Recent works \cite{sarica2020technet, luo2021guiding} offered to retrieve inspirations from patent data, but focused on semantic similarity, not structural similarity or functional relations.
\citet{murphy2014function} did try to encode some functionality information, but this was based on shallow keyword embedding, without taking abstraction into account. 

More directly relevant to functional thinking are approaches that explicitly encode functional knowledge. The Functional Basis \cite{hirtz2002functional, stone1999development}  provides a standardized vocabulary for describing the functions and flows within a system. This framework has been used to develop tools for concept generation, but the vocabularies are often small and restricted, and do not offer the expressivity of our approach.



While analogy and abstraction are still considered hard tasks for machines \cite{mitchell2021abstraction}, LLMs have shown emergent capabilities of analogical reasoning \cite{webb2023emergent, zhou2025self}. This raises exciting possibilities for future work.

\section{Conclusions and future work}

In this work we propose a method to build Functional Concept Graphs --  representations that enable navigation across interconnected functional elements, 
facilitating abstraction and reframing of problems, 
and the discovery of analogical inspirations.
Unlike previous attempts, our approach can scale to large datasets and results in  richer, better-connected and less noisy graphs, whose edges explicitly encode abstraction relations.  
We also introduce MUSE, an algorithm that, given an FCG and a target problem, produces inspirations that could help users creatively solve the problem.

We demonstrate our method by computing an FCG on 500K patents, which we release for further research.
%
%
We conduct a user study to evaluate the usefulness of MUSE. Our results indicate that our inspirations resulted in more creative ideas, both in terms of absolute number (up to 19\% more creative solutions when using inspirations) and ratio (49\% creative ideas without inspirations opposed to 75\% in the sentence condition).

In this work, we suggested a simple way of sampling inspirations from the graph. In the future, we plan to explore more sophisticated sampling methods. One immediate option is to sample far-off analogies as inspirations. 

Another interesting research direction is using the inspiration graph to enhance creativity in SOTA AI agents. Our initial inspection in Section \ref{sec:results:insights} demonstrates that current SOTA LLMs struggle generating truly never-before-seen solutions. We hypothesize that enriching these models in either the training or inference phases would help enhancing their creative problem-solving ability. 

We hope our work would inspire further research on enhancing creative ideation by automatically finding structured representations for navigating the design space and finding analogies in large, complex idea repositories.
\section{Ethical considerations}
\xhdr{Experiment} Our used study was approved by an institutional ethics committee. We do not save any personal information for any of the participants apart from the Prolific ID, which we discard after completing the analysis. Ideas generated by the subjects remain their own, and we make no commercial use of any of them. Prior to starting the experiment, participants agreed to privacy and data collection terms which were fully described in a consent form.

\xhdr{Usage of AI agents} We did not use any AI agents the writing process of this paper. For coding, we occasionally used Claude 3.7 Sonnet, and verified the output code. The parts of our pipeline that include the usage of LLMs were explicitly described in Section \ref{sec:methods}.

\xhdr{Reliance on automatic creativity} Our work proposes an automatic assistance to enhance creativity, thus reducing the burden of trying to break the creative fixation for the users. In case our method becomes popular, one might rely on it in creative problem-solving tasks. This raises the potential risk of over-reliance on automatic creativity tools and creating biases.

\section{Limitations}
In this work, we opted to use patents extracted from the US patents database, and use these patents to draw inspirations from existing solutions. One limitation of our method is that we might miss out on many relevant products and ideas that do not exist in the database. Specifically, users using our solution cannot be inspired by patents not written in English and registered in the US, as they are not part of our dataset. This might introduce a  bias and fixate the users to draw analogies from certain types of solutions, ignoring solutions from different cultures.

Similarly, all participants in our user-based study were native English speakers. We did not test how our tool helps ideation for non-native English speakers. 

Another limitation of our work is the reliance on CPC tags to collect the mechanism tags. When generalizing to new domains, we would need alternative methods for collecting mechanism tags.


\bibliography{custom}

\appendix

\section{Annotated patent example}
\label{subsec:appendix: annotated patent}
\begin{figure}[h]
\includegraphics[scale = 0.27]
{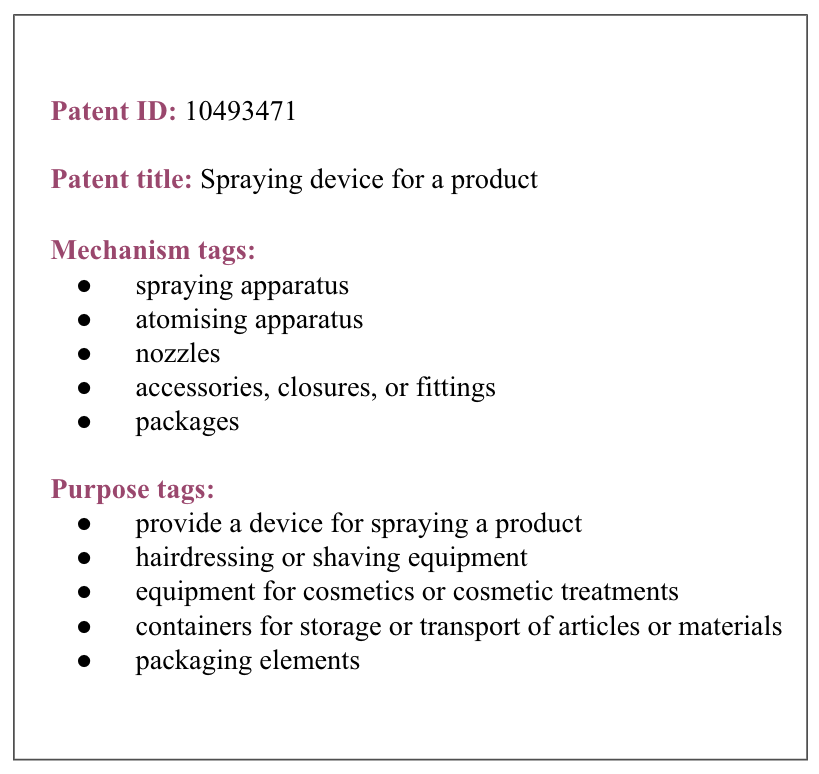}
\caption{Example of an annotated patent. The patent title is associated with purpose and mechanism tags, demonstrating the problems and solutions offered in it.}
\label{fig:appendix:annotated_data}
\end{figure}

See figure \ref{fig:appendix:annotated_data} for a full example of an annotated patent.

\section{Annotation prompt example}
\label{sec:appendix: annotation prompt}
See figure \ref{fig:appendix:prompt} for a prompt and output example for annotating 

\begin{figure*}[t]
\centering
\fbox{%
\begin{minipage}{0.99\textwidth}
\textbf{Prompt input and output example for annotating a product description} \\ \\
\textbf{Prompt:} \\ \\
\textbf{Example 1:}\\
Method and device for precise invasive procedures.This invention relates generally to the field of invasive medical procedures, and specifically to accurate monitoring of invasive procedures with an imaging system . A method for inserting an invasive tool, including: attaching a frame to a human body adjacent to a portion of the body; acquiring an image of the body; determining a trajectory of the tool on the image; calculating points of intersection between the trajectory and two sheet which are adapted to be inserted into the frame; perforating the sheets at the calculated points; placing the sheet within the frame; and inserting the invasive tool through the perforations. \\
\textbf{The purpose of the patent is to monitor invasive procedures.} \\

\textbf{Example 2:}\\
Hair style device. This invention relates to devices which attach to the hand of a user ,which devices simultaneously blow-dry and style hair on mammals. A hair-styling device containing a hand attachment and a source for heated air under pressure connected to the hand attachment.\\
\textbf{The purpose of the patent is to blow dry and style hair} \\

\textbf{Example 3:} \\
Swivel wheel mount. This invention relates to a child's stroller (a bassinet , a baby buggy or similar device used to support or transport a person) with wheels which swivel. Disclosed is a stroller including a frame member, a swivel mount adapted to receive the frame member, a swivel latch adapted to be received in the swivel mount, a suspension housing including a swivel latch receiving portion, the suspension housing adapted to be attached to the swivel mount, a swivel pin adapted to be received in the frame member, the swivel mount, and the suspension housing, and at least one wheel pivotally attached to the frame member.\\
\textbf{The purpose of the patent is to provide a stroller with a swivel 
wheel} \\

\textbf{Your input:}\\
Modular engine, such as a jet engine, with a speed reduction gear. The present invention relates to an aircraft propulsion engine, such as a turbojet engine, a multi-flow turbofan, in particular with a high dilution ratio, or a turboprop engine, having a front power transmission shaft, driven by a turbine rotor by means of a speed reduction gear. The present invention relates to an engine (1) with a modular structure comprising a plurality of coaxial modules (A, B, C) with, at one end, a first module (A) comprising a power transmission shaft (3) and a speed reduction gear (7), said power transmission shaft being driven via the speed reduction gear (7) by a turbine shaft (2) secured to one (C) of said coaxial modules that is separate from the first module, the speed reduction gear comprising a drive means (8 and 9) fixed to the turbine shaft (2) and to a journal (13) of a shaft of a low-pressure compressor rotor (1 a), characterized in that it comprises a first nut (16) for fastening the drive means to the journal and a second nut (14) for fastening the drive means to the turbine shaft.\\
\textbf{What is the purpose of the patent? What is the context of the patent?}\\ \\

\textbf{Output example:}\\
\textbf{The purpose of the patent is to provide a method for reducing the speed of a jet engine.}

\end{minipage}
}
\caption{An example of the prompt we use to annotate patent descriptions with purpose tags, followed by an output example. The in-context prompt consists of 3 examples for annotated patent descriptions, followed by the description to be tagged.}
\label{fig:appendix:prompt}
\end{figure*}

\section{Implementation details}
\label{sec:appendix:implementation details}
\subsection{General implementation details}
All code for this project was written in python-3.10. The attached Git repository contains all code, dependencies and commands required to create the inspiration graph, sampling from it, running the experiment and analyzing the results. All packages and datasets used in this work were used solely for academic work, with accordance to their license. All data statistics, model and hyper-parameter choices are described in their corresponding sections. To create the inspiration graph, we used a single A5000 GPU (24GB) for ~4 hours. We mostly used it to run the NLI  and llama3.1-8b-Instruct models described in section \ref{sec: methods: building graph}. In order to speed up the agglomerative clustering process described in section~\ref{sec:method:purpose nodes}, we used 8 32GB-CPUs that ran in parallel for ~3 days. 

\subsection{Getting mechanism tags from CPC tags}
\label{sec:appendix:implementation details: mechanism}
Prior to training the mechanism classifier described in section~\ref{sec:methods:getting annotations}, we process each one of the CPC titles text. first, if it contains more than one text span, we split it so each CPC id may be indicated by multiple titles in our final tags set. This split to multiple text spans may result in a single patent having multiple tags, which are not all necessarily related to it. To deal with this, for each patent we measure the cosine similarity between its title and the CPC tags title's Sentence BERT embeddings, and select only the most relevant one. 

\subsection{Clustering implementation details}
\label{sec:appendix:implementation details: clustering parameters}
For all clustering purposes, we used the algorithms implemented in Scikit-learn \cite{scikit-learn} used under BSD License.
\xhdr{Loose clustering with K-means} As explained in section \ref{sec: methods: building graph}, the loose clustering step is meant to reduce the number of nodes which we aggressively cluster with agglomerative clustering. We choose to use K-means with $K=10000$. 

\begin{figure}[t]
    \centering
    \includegraphics[width=0.85\textwidth]{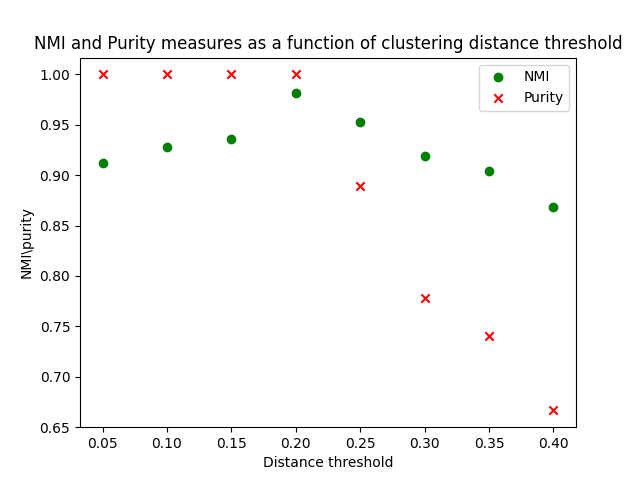}
    \caption{Purity (green dots) and NMI (red X) results for the different similarity thresholds. We see that the best results for both metrics are produced by choosing distance threshold $=0.2$.}
    \label{fig:nmi_purity}
\end{figure}

\xhdr{Aggressive clustering} In order to select the parameters for the agglomerative clustering step, we rub agglomerative clustering on the evaluation dataset mentioned in section \ref{sec:method:purpose nodes} with similarity thresholds $\{i \cdot 0.05 \mid i \in \{1, 2, \ldots, 8\}\}$.  We use cosine similarity as the metric for clustering, complete linkage and the default values for all other parameters. Figure \ref{fig:nmi_purity} shows the NMI and purity measures across the tested distance thresholds.

\subsection{Abstraction with entailment} 
\label{sec:appendix:implementation details: entailment}
In order to check whether a problem node entails  another, we randomly select a representative purpose tag from each node. We experiment with 2 prefix options -- ``I want'' and ``The patent provides''. For the NLI model, we use a fine-tuned version of Deberta-V3-large (304M parameters) \cite{he2021debertav3}, offered in \citet{laurer2024less}. In order to choose the entailment threshold, we tag the clusters generated in section \ref{sec:appendix:implementation details: clustering parameters}, and tag all abstraction relations between them.  We test different values for the entailment threshold $t$ and prefix, settling on $t=0.5$ and prefix = ``I want''  since it achieved the highest recall (0.65) and precision (0.9) rates.

\subsection{Enhancing connectivity with LLM-based connections and verb-based connections}
We extracted verbs using NLTK \cite{bird2006nltk}, used under Apache License Version 2.0.
\label{sec:appendix: implementation details: LLMs}
As we mentioned in section \ref{sec: methods: building graph}, we enhance the connectivity of the graph by creating additional LLM-based nodes created from a set of candidate nodes. We first discuss the process of choosing these nodes. Let $h_max$ be the maximal height of a node in the loose cluster graph. We choose candidate nodes as all nodes whose height is at least $h_{max}- 3$ and distance from the highest node in their path is 2. After selecting the candidate nodes, we use K-means with $K = 5$ to split these nodes into 5 clusters. We prompt llama3.1-8b-instruct to select (if possible) a subset of the nodes from each cluster that can be abstracted together, and find an abstraction for them. As an additional validation step, for each such cluster $c_i$, we find the furthest cluster $c_j$ by computing  the minimum cosine similarity between all nodes. We add to $c_i$ to 2 nodes which are most dissimilar to nodes from $c_j$. If any of these nodes were selected alongside original nodes from $c_i$ during the abstraction process, we discard the abstraction.

\subsection{Connecting the interim graphs}
\label{sec:appendix: implementation details: interim graph}
Similar to the process of enhancing the graph connectivity using LLM-based nodes described in section~\ref{sec: methods: building graph}, in order to connect the different interim graph we first select a set of candidate nodes for each loose cluster. We select the same candidate nodes as those described in section~\ref{sec:appendix: implementation details: LLMs}. In order to connect these nodes with NLI-based nodes, we perform the same process described in section \ref{sec: methods: building graph}, over pairs of nodes coming from different candidate node sets. For the LLM-based enhancements, we replicate the same process described in section~\ref{sec:appendix: implementation details: LLMs}.

\section{Inspiration examples}
\label{sec: appendix: inspiration examples}
Figure \ref{fig:appendix:inspiration exmamples} shows an example of 5 inspirations presented in each condition in our experiment.

\begin{figure*}[t]
\centering
\fbox{%
\begin{minipage}{0.99\textwidth}
\small  
\textbf{Examples for inspirations provided in the experiment}

\textbf{Condition 1: Purpose}

\begin{enumerate}
\item Possible inspiration: Think of \textbf{a method and apparatus for cooling a work piece}
\item Possible inspiration: Think of \textbf{a system for cooling a person}
\item Possible inspiration: Think of \textbf{a water cooled door}
\item Possible inspiration: Think of \textbf{a computer cooling assembly}
\item Possible inspiration: Think of \textbf{a cooling bed system}
\end{enumerate}

\textbf{Condition 2: Purpose + Mechanism}

\begin{enumerate}
\item Possible inspiration: Think of \textbf{a method and apparatus for cooling a work piece}
Related concepts: 
\begin{itemize}
    \item Heat-exchange apparatus
\end{itemize}
\item Possible inspiration: Think of \textbf{a system for cooling a person}
Related concepts:
\begin{itemize}
    \item Air-humidification
\end{itemize}
\item Possible inspiration: Think of \textbf{a water cooled door}
Related concepts:
\begin{itemize}
    \item Combustion engines
\end{itemize}
\item Possible inspiration: Think of \textbf{a computer cooling assembly}
Related concepts:
\begin{itemize}
    \item Vehicle cooling systems
\end{itemize}
\item Possible inspiration: Think of \textbf{a cooling bed system}
Related concepts:
\begin{itemize}
    \item Therapeutic cooling beds
\end{itemize}
\end{enumerate}

\textbf{Condition 3: Purpose + Mechanism sentence}

\begin{enumerate}
\item Possible inspiration: Think of \textbf{a method and apparatus for cooling a work piece}
Related concepts: 
\begin{itemize}
    \item Heat-exchange apparatus without direct contact enables precise workpiece cooling
\end{itemize}
\item Possible inspiration: Think of \textbf{a system for cooling a person}
Related concepts:
\begin{itemize}
    \item Air-humidification enhances evaporative cooling effects for personal comfort
\end{itemize}
\item Possible inspiration: Think of \textbf{a water cooled door}
Related concepts:
\begin{itemize}
    \item Combustion engines employ water cooling technologies for component protection
\end{itemize}
\item Possible inspiration: Think of \textbf{a computer cooling assembly}
Related concepts:
\begin{itemize}
    \item Vehicle cooling systems inform compact computer cooling assembly design
\end{itemize}
\item Possible inspiration: Think of \textbf{a cooling bed system}
Related concepts:
\begin{itemize}
    \item Medical science applications incorporate therapeutic cooling beds for patient care
\end{itemize}
\end{enumerate} 

\end{minipage}
}
\caption{Examples for inspirations sampled for the problem ``Cool a room''. We provide 5 examples for each condition. For clarity, we show the same problem and solution nodes sampled in each condition.}
\label{fig:appendix:inspiration exmamples}
\end{figure*}

\section{Experiment instructions}
\label{sec:appendix:expriment instructions}
\begin{figure*}
    \centering
    \begin{subfigure}[b]{\textwidth}
        \centering
        \includegraphics[width=\textwidth]{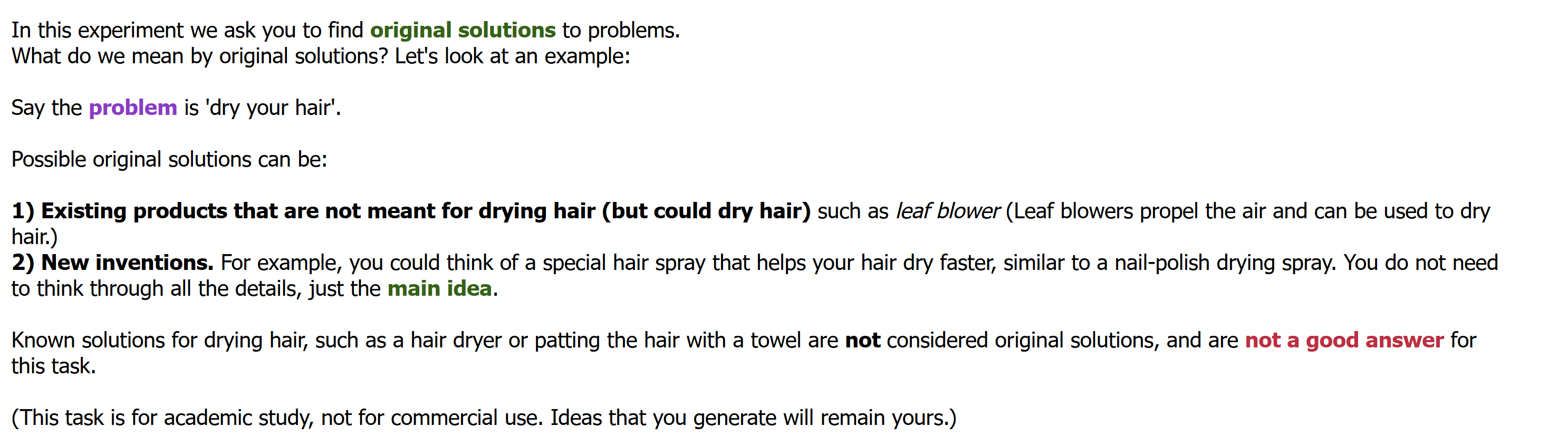}
        \label{fig:instructions:subfig1}
    \end{subfigure}
    \hfill
    \begin{subfigure}[b]{\textwidth}
        \centering
        \includegraphics[width=\textwidth]{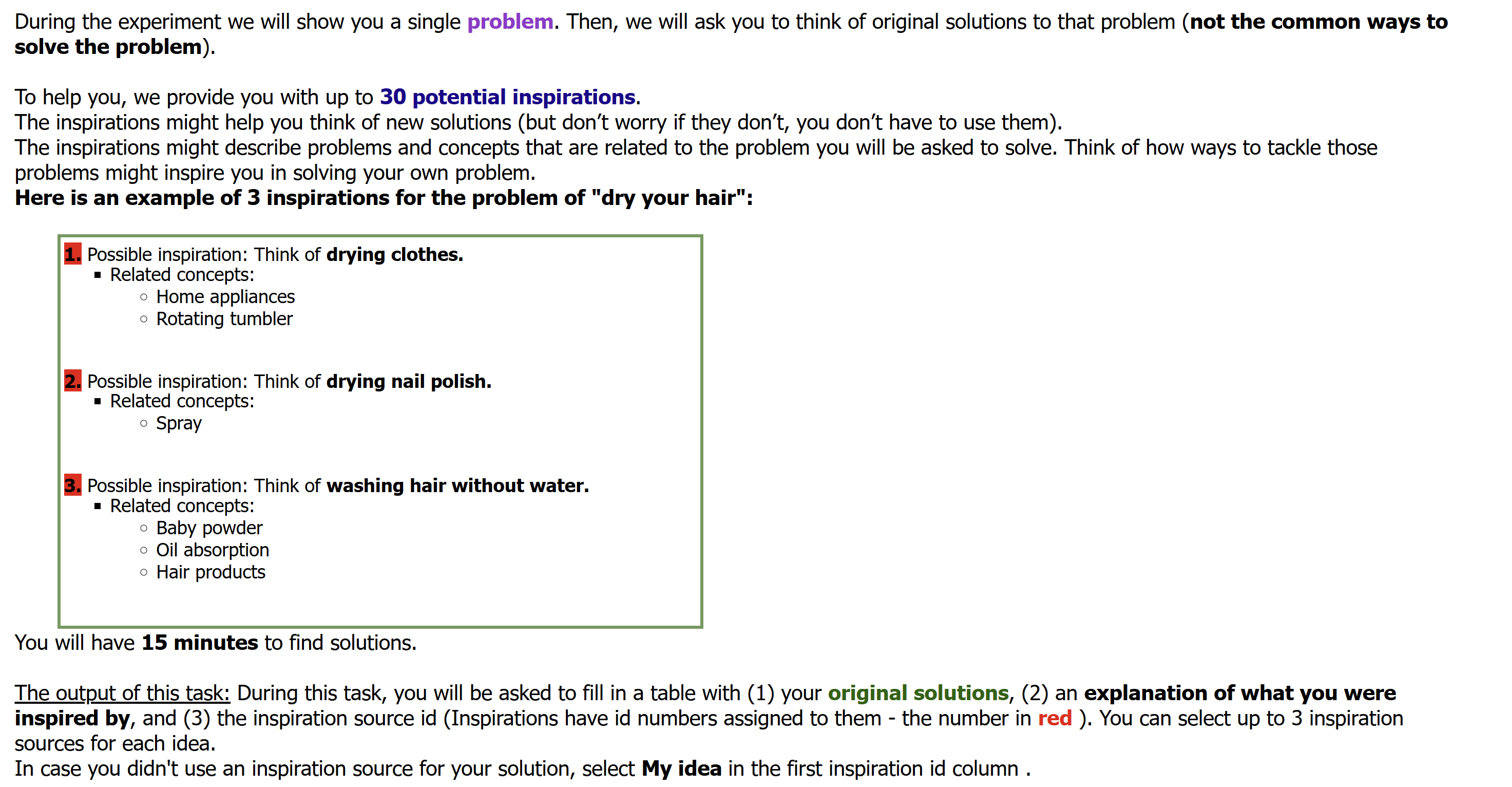}
        \label{fig:instructions:subfig2}
    \end{subfigure}
    \begin{subfigure}[b]{\textwidth}
        \centering
        \includegraphics[width=\textwidth]{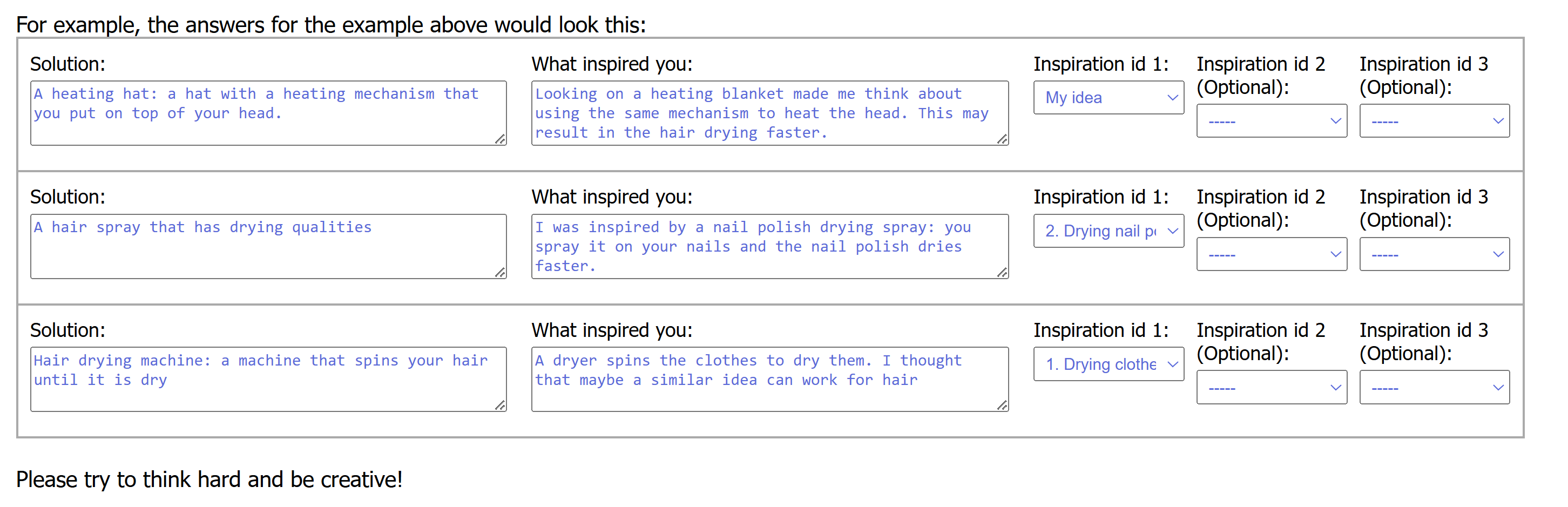}
        \label{fig:instructions:subfig3}
    \end{subfigure}
    \caption{The full instructions for our experiment }
    \label{fig:appendix:experiment instructions}
\end{figure*}

The full instructions for our experiment are provided in figure \ref{fig:appendix:experiment instructions}.

\end{document}